\documentclass[journal]{IEEEtran}
\usepackage{graphicx}
\usepackage{multirow}
\usepackage[cmex10]{amsmath}
\usepackage[tight,footnotesize]{subfigure}

\begin{document}

\title{Beyond $\chi^2$ Difference: Learning Optimal Metric for Boundary Detection}

\author{Fei~He~and~Shengjin~Wang,~\IEEEmembership{Member,~IEEE}%
\thanks{This work was Supported by the National High Technology Research
and Development Program of China (863 program) under Grant No. 2012AA011004
and the National Science and Technology Support Program under Grant No. 2013BAK02B04.}%
\thanks{The authors are with the Department of Electronic Engineering,
Tsinghua University, Beijing 100086, China
(e-mail: hef05@mails.tsinghua.edu.cn; wgsgj@tsinghua.edu.cn).}}

\maketitle

\begin{abstract}
This letter focuses on solving the challenging problem
of detecting natural image boundaries.
A boundary usually refers to the border
between two regions with different semantic meanings.
Therefore, a measurement of dissimilarity
between image regions plays a pivotal role
in boundary detection of natural images.
To improve the performance of boundary detection,
a \emph{Learning-based Boundary Metric (LBM)} is proposed
to replace $\chi^2$ difference adopted
by the classical algorithm \emph{mPb}.
Compared with $\chi^2$ difference, LBM is composed
of a single layer neural network and an RBF kernel,
and is fine-tuned by supervised learning rather than human-crafted.
It is more effective in describing the dissimilarity
between natural image regions while tolerating large variance of image data.
After substituting $\chi^2$ difference with LBM,
the F-measure metric of \emph{mPb} on the BSDS500 benchmark
is increased from 0.69 to 0.71.
Moreover, when image features are computed on a single scale,
the proposed LBM algorithm still achieves
competitive results compared with \emph{mPb},
which makes use of multi-scale image features.
\end{abstract}


\section{Introduction}
\label{sec:intro}

\IEEEPARstart{N}{atural} image boundary detection is a fundamental problem
in the field of image processing and computer vision.
The boundaries can be used as low-level image features
for object classification and detection
\cite{opelt2006, shotton2008, farhadi2009, ferrari2010}.
For example, the algorithm proposed by \cite{opelt2006} detects cows and horses
by matching boundary fragments extracted from images.
In this case, clean boundary maps are required for follow-up stages.
Due to the ambiguity of low-level features
and the lack of semantic information,
boundary detection remains a challenging problem
after decades of active research
\cite{canny1986, dollar2006, kokkinos2010, kennedy2011}.
This letter proposes a \emph{Learning-based Boundary Metric (LBM)}
and makes efforts to improve the performance of a classical algorithm
named \emph{Multi-scale Probability of Boundary (mPb)} \cite{arbelaez2011}.

A boundary usually refers to the border
between two regions with different semantic meanings.
Therefore, measuring the dissimilarity between image regions
is at the core of boundary detection.
In a canonical framework, we first extract local image features,
such as brightness histogram, from an image.
Then the distance of descriptors from adjacent regions
is used as an indicator to boundary response.
With a good measurement, the boundary response should
be weak inside a sematic region while strong on the border.

To find an ideal measurement,
both feature extraction and distance calculation are crucial.
Earlier researchers prefer relatively simple features and metrics
due to limited computing resources.
For example, Canny detector introduced by \cite{canny1986}
uses analytic derivatives of brightness cue
to compute boundary response.
However, brightness discontinuity exists
not only on borders between different regions
but also inside a semantic region.
The Canny detection results usually contain lots of non-boundary points.
A later algorithm named \emph{Probability of Boundary (Pb)} \cite{martin2004}
suggests combining multiple cues for boundary detection.
It proposes a histogram-based feature
to fully exploit brightness, color and texture cues.
Furthermore, $\chi^2$ difference is adopted to calculate the distance,
since it is shown to be more effective in the histogram-based feature space.
With the new feature and $\chi^2$ difference,
\emph{Pb} is capable of detecting complex boundaries
while eliminating most noise, making a big step forward.
\emph{Multi-scale Probability of Boundary (mPb)}
proposed by \cite{arbelaez2011} is the successor of \emph{Pb}.
Compared with the predecessor,
\emph{mPb} computes the features on multiple scales.
As shown in experiments of \cite{ren2008},
multi-scale cues improve the performance of boundary detection.

\begin{figure}[!t]
\centering
\includegraphics[width=7.2cm]{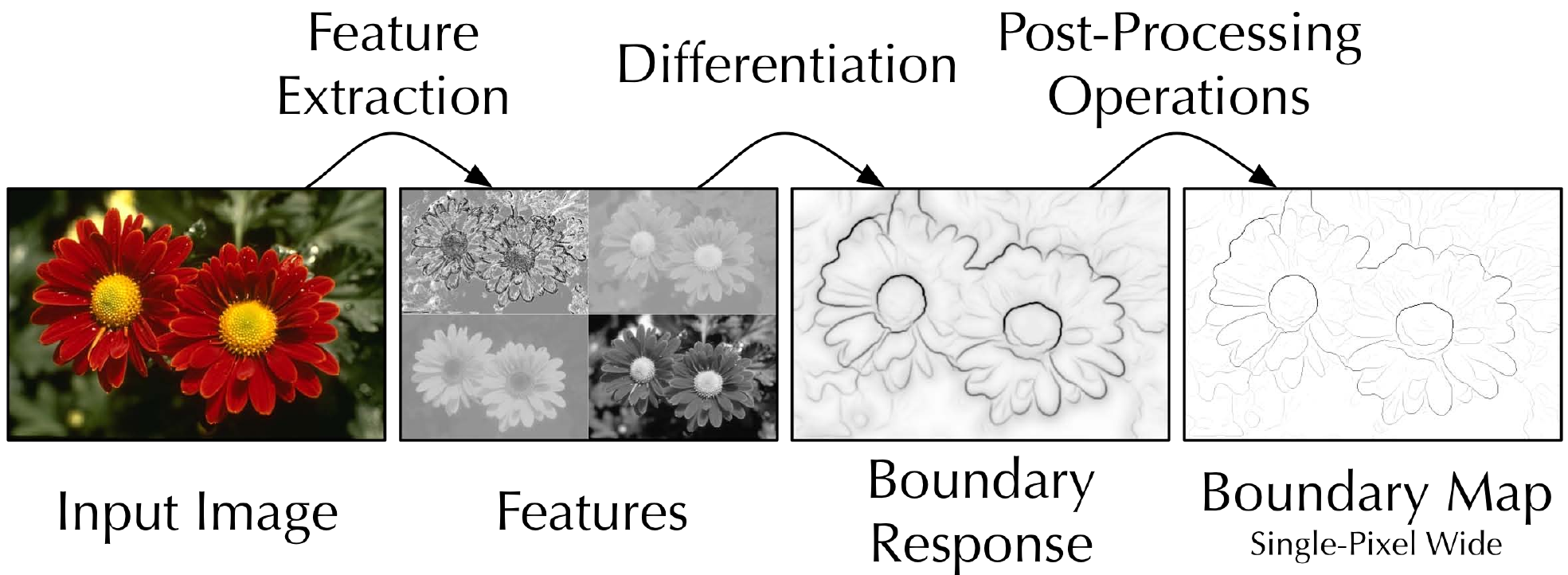}
\caption{A canonical framework of boundary detection.
The first step converts the input image into the feature space.
For \emph{mPb}, the features have 4 channels,
including 3 channels of \emph{Lab} color space and 1 channel of textons.
Then descriptor distances are calculated in the second step.
In the end, a single-pixel wide boundary map
is generated after the post-processing operations.}
\label{fig:framework}
\end{figure}

For both \emph{Pb} and \emph{mPb},
one of the highlights is to learn parameters
from human annotations in dataset BSDS300 \cite{martin2001}.
By introducing a learning stage,
researchers hope to capture the implicit structure
of natural image data and further improve the performance.
However, the drawback of human-crafted metrics
such as the $\chi^2$ difference consist in their limited fitness to the data.
In fact, experiments in this letter show that
the improvement brought by supervised learning is relatively minor.
Inspired by \cite{eric2003}, we propose to learn a distance metric
to substitute the $\chi^2$ difference in \emph{mPb}.
Different from \cite{eric2003}, the \emph{Learning-based Boundary Metric (LBM)}
is composed of a single layer neural network and an RBF kernel,
and is fine-tuned by strongly supervised learning.
After applying LBM, the F-measure metric of \emph{mPb}
on the BSDS500 benchmark is increased from 0.69 to 0.71.
The following parts will show details of LBM
and evaluation results on BSDS500 \cite{arbelaez2011}.

\section{Learning-based Boundary Metric (LBM)}
\label{sec:lbm}

A canonical framework of boundary detection
typically consists of three steps,
i.e., feature extraction, differentiation and post-processing operations,
as illustrated in Fig.~\ref{fig:framework}.
Taking \emph{mPb} for an example,
histograms of different cues and scales are firstly extracted.
Then, the distance of descriptors
from adjacent regions is calculated using $\chi^2$ difference.
Finally, post-processing operations,
such as noise reduction, cues fusion and oriented non-maximum suppression,
are employed to generate single-pixel wide boundary maps as the output.

\subsection{Histogram-based Feature and $\chi^2$ Difference}
\label{sec:lbm:chi}

\begin{figure}[!t]
\centering
\includegraphics[width=6.4cm]{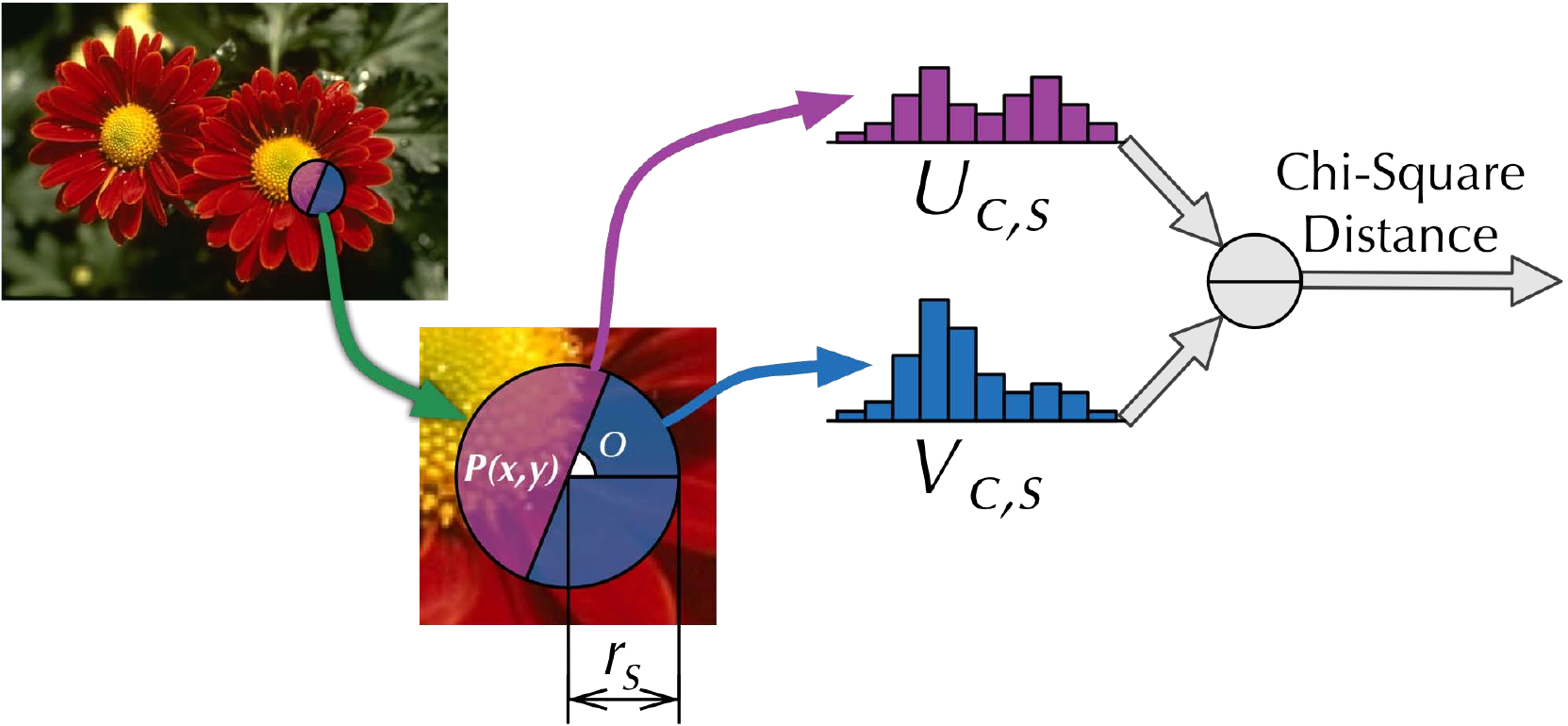}
\caption{Histogram-based feature of \emph{mPb}.
Given a pixel $P(x,y)$ and the orientation $o\in[0,\pi)$,
a pair of features is extracted, denoted by $U_{c,s}$ and $V_{c,s}$.
They are both histograms which represent the distribution
of cue $c$ within a half disk at scale $s$.
The value of $r_s$ depends on at which scale feature vectors are extracted.
After feature extraction, $\chi^2$ difference is applied
to calculate distance between $U_{c,s}$ and $V_{c,s}$.}
\label{fig:feature}
\end{figure}

In this letter, we adopt \emph{mPb} \cite{arbelaez2011} as the baseline
and use exactly the same feature.
Given a pixel $P(x,y)$ and the orientation $o\in[0,\pi)$,
feature pairs of different cues and scales
are extracted by pooling pixel-wise features over two half disks.
As shown in Fig.~\ref{fig:feature}, each pair of feature vectors,
$U_{c,s}$ and $V_{c,s}$, corresponds to one kind of cue and a pooling scale.
Both $U_{c,s}$ and $V_{c,s}$ are histograms
which represent the distribution of cue $c$ within a half disk at scale $s$.
Here 4 kinds of cues are considered,
including 3 channels of \emph{Lab} color space and 1 channel of textons.
The number of pooling scales is also 4,
indicating that 16 pairs of feature vectors
are extracted at each pixel and each orientation.

For the traditional approach of $\chi^2$ difference,
each pair of feature vectors can be used to compute a distance $d_{c,s}$,
\begin{equation}
\label{eq:chidist}
d_{c,s}=\frac{1}{2}\sum_{m}{\frac{(U_{c,s,m}-V_{c,s,m})^2}{U_{c,s,m}+V_{c,s,m}}} \; .
\end{equation}
Then, all the distances computed in Eq.~\ref{eq:chidist}
are collected and summed up with respect to $c$ and $s$,
weighted by $w_{c,s}$ obtained from logistic learning,
\begin{equation}
\label{eq:chisum}
d=\sum_{c=0}^{3}{\sum_{s=0}^{3}{w_{c,s}d_{c,s}}} \; .
\end{equation}
The result $d$ characterizes the boundary strength
at pixel $P$ and orientation $o$.
The pipeline of \emph{mPb} is illustrated in Fig.~\ref{fig:pipeline:chi}.

The $\chi^2$ difference approach of \emph{mPb} has a shortcoming
in which supervising information affects only the weights $w_{c,s}$,
while most parts of the algorithm are human-crafted.
Restricted by the number of tunable parameters,
the algorithm cannot fit the image data very well.
In fact, if distances $d_{c,s}$ are summed up with equal weights,
the F-measure metric on BSDS500 remains almost the same.
Table~\ref{table:mpb} demonstrates the results of \emph{mPb}
with both learned weights and equal weights.
ODS or OIS in the table refers to the best F-measure
for the entire dataset or per image respectively,
and AP (Average Precision) is the area under the PR curve.
Details of evaluation method can be found in Section~\ref{sec:eval}.

\begin{table}[!t]
\caption{Evaluation Results of \emph{mPb}}
\label{table:mpb}
\centering
\begin{tabular}{|c|c|c|c|}
\hline
{ } & ODS & OIS & AP\\
\hline
\emph{mPb} (with learned weights) & 0.69 & 0.71 & 0.68\\
\hline
\emph{mPb} (with equal weights) & 0.69 & 0.71 & 0.70\\
\hline
\end{tabular}
\end{table}

\subsection{Learning Optimal Boundary Metric}
\label{sec:lbm:lbm}

According to the aforementioned analysis,
the learning stage of \emph{mPb} achieves limited improvements.
To obtain better results, it is necessary
to increase the number of tunable parameters.
In this section, boundary metric is introduced,
which is then optimized with respect to
the loss function defined by Eq.~\ref{eq:loss}.

As is known, \emph{Artificial Neural Network (ANN)}
is widely recognized for its strong fitting capability.
Accordingly, the proposed LBM builds a neural network for each cue and scale
to transform the local features into a new space.
Then the distance of features is computed in the transformed space.
In this manner, supervising information can be used
to learn a better space where the metric is more consistent with human annotations.
Assuming $f_{c,s}(\cdot)$ is the transformation
corresponding to cue $c$ and scale $s$,
the new distance can be formatted as follows,
\begin{equation}
\label{eq:lbmdef}
d_{c,s}^{LBM}=D(\widetilde{U}_{c,s},\widetilde{V}_{c,s})=D(f_{c,s}(U_{c,s}),f_{c,s}(V_{c,s})) \; ,
\end{equation}
where $D(\cdot,\cdot)$ is the metric of the learned space.
In this letter, we propose to use a group of logistic functions
to implement the transformation,
\begin{equation}
\label{eq:logdef}
\begin{split}
\widetilde{U}_{n}&=f_{n}(U)=\frac{1}{1+e^{-\alpha_{n}-\sum_{m=1}^{M}{\beta_{n,m}U_{m}}}}, \\
\widetilde{U}&=[\widetilde{U}_1,\widetilde{U}_2,\cdots,\widetilde{U}_N]^T \; .
\end{split}
\end{equation}
$M$ and $N$ in the formula denote the dimensions
of input and output features, respectively.
After the transformation, RBF kernel rather than linear kernel
is adopted to compute the distance,
because nonlinear kernel is more suitable
for complex data such as natural images,
\begin{equation}
\label{eq:rbf}
D(\widetilde{U},\widetilde{V})=1-e^{-\frac{\sum_{n=1}^{N}{(\tilde{U}_n-\tilde{V}_n)^2}}{2\sigma^2}} \; .
\end{equation}

\begin{figure}[!t]
\centering
\subfigure[$\chi^2$ difference approach]{
\label{fig:pipeline:chi}
\includegraphics[width=4.2cm]{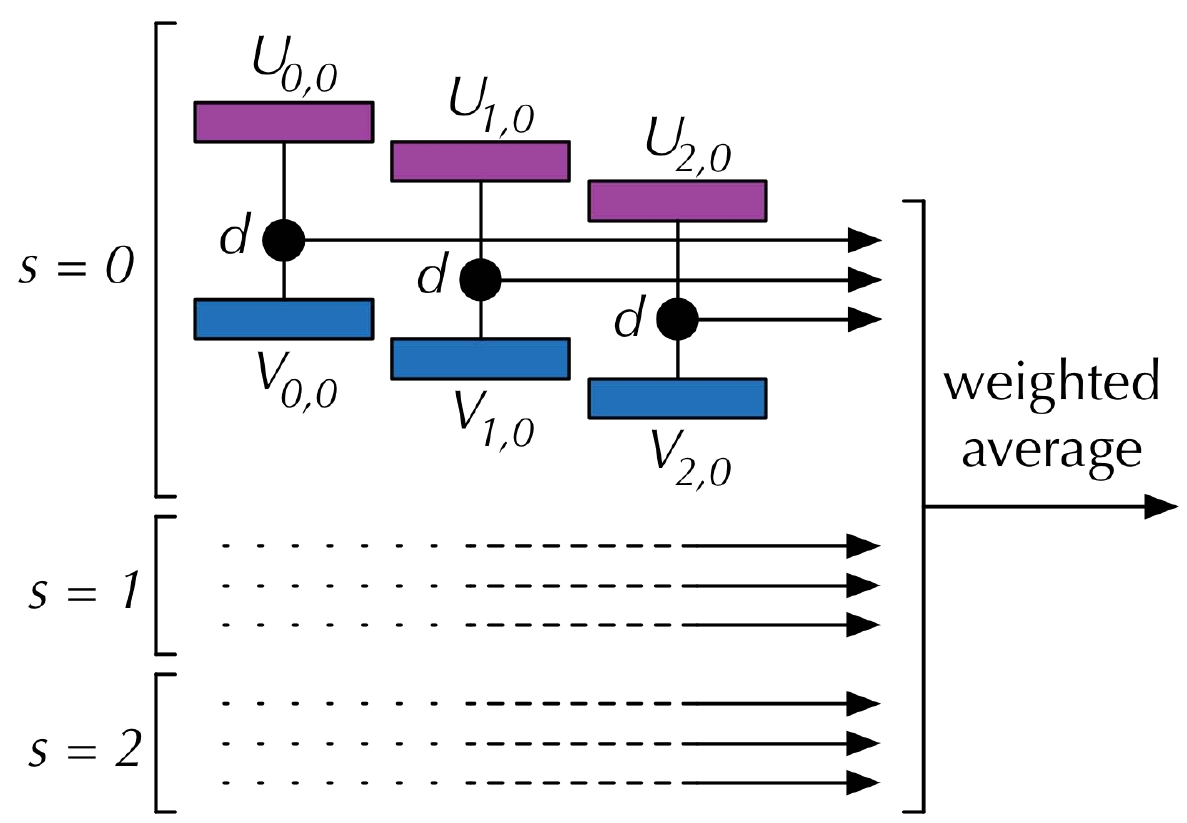}}
\subfigure[LBM approach]{
\label{fig:pipeline:lbm}
\includegraphics[width=4.2cm]{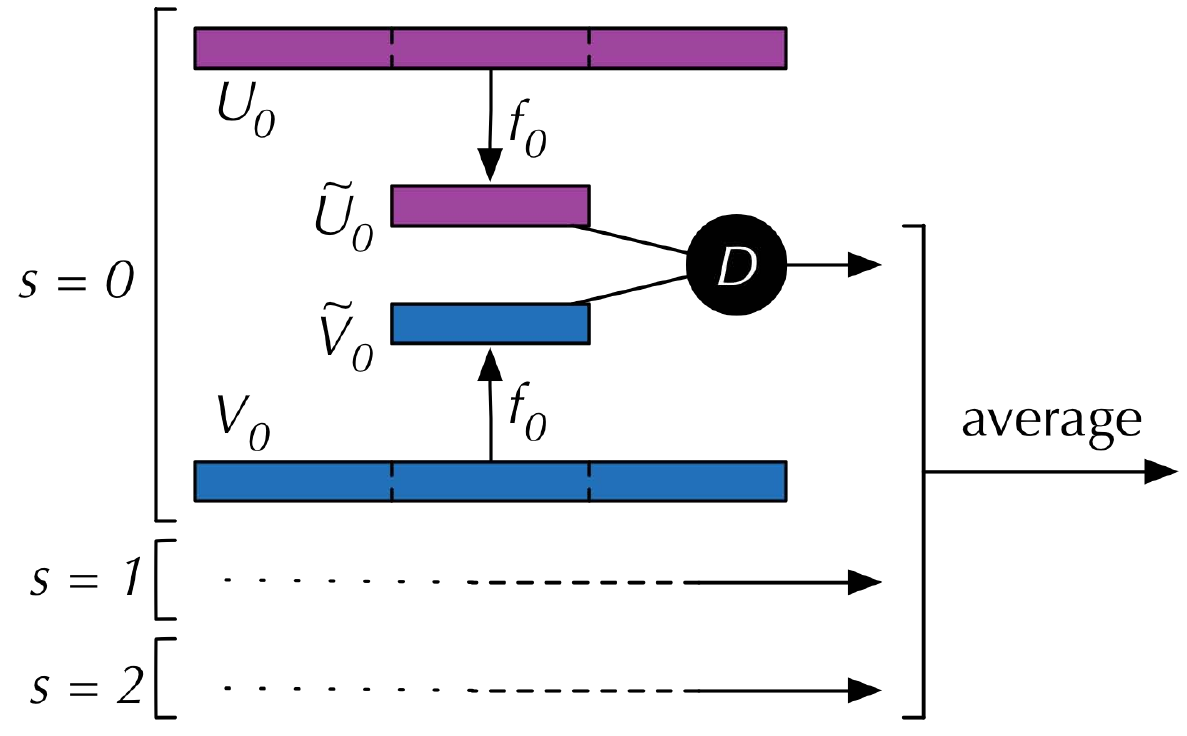}}
\caption{Pipelines of (a) the baseline and (b) the proposed LBM method.
In (a) distances are calculated between all pairs
of feature vectors $U_{c,s}$ and $V_{c,s}$.
The black nodes in the figure denote the $\chi^2$ difference.
The final output is the weighted average
of distances of all cues at all scales.
In (b) the LBM method concatenates feature vectors
of the same scale into a single vector $U_s$.
Then the learning-based transformation $f_s$ is applied on $U_s$ and $V_s$.
At each scale, metric $D$ based on RBF kernel, shown as the black node,
is used to compute distances between the transformed features
$\widetilde{U}_s$ and $\widetilde{V}_s$.
The final output is the average of distances at all scales.
The difference between the two approaches
lies in the mechanism of metric design,
either (a) human-crafted or (b) fine-tuned by supervised learning.}
\label{fig:pipeline}
\end{figure}

Until now, we have introduced the basic structure of LBM.
In the final implementation, feature vectors of the same scale
are concatenated to form a single vector,
allowing more interactions among different cues.
Then, a larger neural network is learned for
$U_s=[U_{0,s}^T,U_{1,s}^T,U_{2,s}^T,U_{3,s}^T ]^T$.
In the end, the mean of descriptor distances at all scales,
$d^{LBM}$, is computed as output of the boundary response,
\begin{equation}
\label{eq:lbmsum}
\begin{split}
d_{s}^{LBM}&=D(\widetilde{U}_s,\widetilde{V}_s)=D(f_s(U_s),f_s(V_s)), \\
d^{LBM}&=\frac{1}{4}\sum_{s=0}^{3}{d_{s}^{LBM}} \; .
\end{split}
\end{equation}
The pipeline of LBM is illustrated in Fig.~\ref{fig:pipeline:lbm}
as a comparison with the \emph{mPb} approach.

With the above definitions, the next step is
to learn parameters $\alpha_n$ and $\beta_{n,m}$
according to human annotations.
We define a loss function to indicate
how well the neural networks fit the data,
and then use \emph{Stochastic Gradient Descent (SGD)}
to tune the parameters.
A simple way to define the loss function is directly using $d^{LBM}$,
where losses of boundary and non-boundary pixels
are $1-d^{LBM}$ and $d^{LBM}$ respectively.
However, we prefer the log-style loss function
since the gradient of a non-boundary pixel won't be zero when $d^{LBM}=1$.
In the following definition, $k$ denotes the index of training samples
and $y_k$ is the annotation,
\begin{equation}
\label{eq:loss}
L=-\sum_{k=1}^{K}{y_{k}\log{d^{LBM}}+(1-y_k)\log{(1-d^{LBM})}} \; .
\end{equation}
$y_k=1$ indicates that the $k$th sample is a boundary pixel and vice versa.

To generate training samples,
$\alpha_n$ and $\beta_{n,m}$ are randomly initialized,
sampled uniformly from range $[-1,1]$.
Then the algorithm selects a random image
from the training set to detect boundary pixels with current parameters.
The pixels matched to human annotations are collected as positive training set,
while those without any match are regarded as the negative set.
After that, SGD is performed to update the parameters.
Next, another image is selected and the same process is repeated.
We terminate the learning loop if the F-measure metric
on validating set no longer has a noticeable improvement.
In our implementation, boundary metrics at different scales are learned separately.

\section{Experiments}
\label{sec:eval}

The proposed LBM is evaluated on BSDS500.
The dataset contains 200 testing images,
with about 5 annotations from different persons for each image.
We follow the widely used evaluation measurement proposed by \cite{martin2004},
in which a Precision-Recall (PR) curve is drawn
and the F-measure metric is used for comparison.

A boundary pixel is counted as false alarm
\emph{iff} it does not match any annotation pixels.
Note that it is common that several persons
annotate the same pixel as ground truth,
so the pixel may be counted as recall for several times.
If the input boundary responses are real values rather than binary,
a series of thresholds are utilized to obtain the PR curve.

There are 3 parameters which need to be determined before the learning stage.
The first one is $N$, the dimension of the transformed feature space.
The second one is $\sigma$ in the RBF kernel.
With exhaustive search, we choose $N=16$ and $\sigma=0.2$,
with which the algorithm achieves the best performance on validating set.
The last parameter is learning rate.
Large learning rate results in unstable SGD,
while small learning rate leads to slow convergence.
We set learning rate to 0.0001 as a trade off
between robustness and learning efficiency.
Other parameters, including $\alpha_n$ and $\beta_{n,m}$ in Eq.~\ref{eq:logdef},
are learned from human annotations.
The evaluation results during the learning process indicate that
the F-measure, as well as $\alpha_n$ and $\beta_{n,m}$ in Eq.~\ref{eq:logdef},
converges smoothly after dozens of iterations.

Although the structure of LBM is more complicated
than that of $\chi^2$ difference,
our algorithm requires much less computing resource.
To extract $U_{c,s}$ or $V_{c,s}$ in Fig.~\ref{fig:feature},
the original work needs to perform average pooling
in a high dimensional feature space.
However, dimension of $\widetilde{U}_{c,s}$ or $\widetilde{V}_{c,s}$
in LBM is very low, which means the pooling operation can be accelerated.
Using the same computer with Intel i7-2600 and 16GB RAM to test both algorithms,
LBM is able to achieve a $5\times$ speed-up.

\begin{figure}[!t]
\centering
\subfigure[]{
\label{fig:pr:m}
\includegraphics[width=2.7cm]{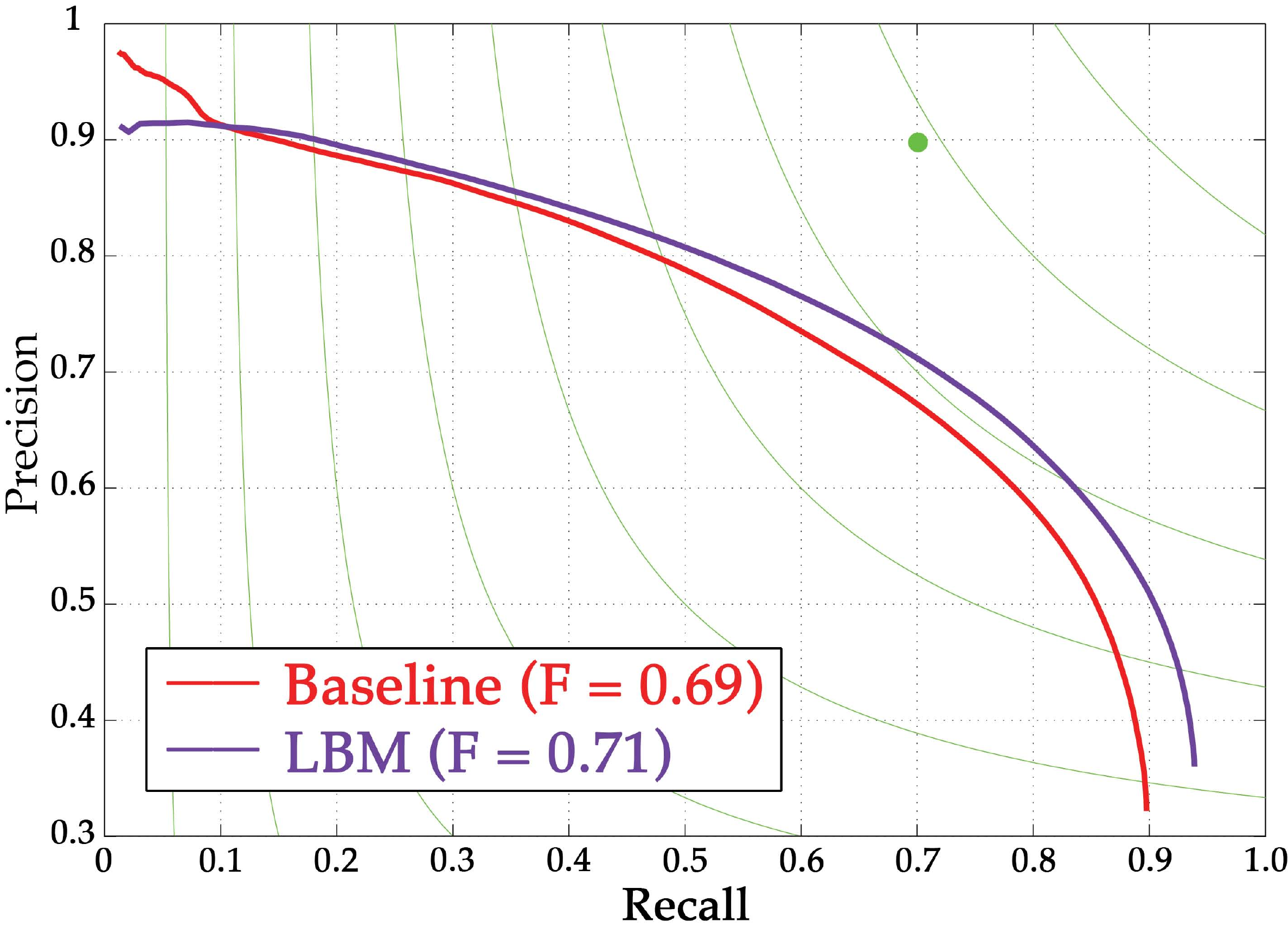}}
\subfigure[]{
\label{fig:pr:s}
\includegraphics[width=2.7cm]{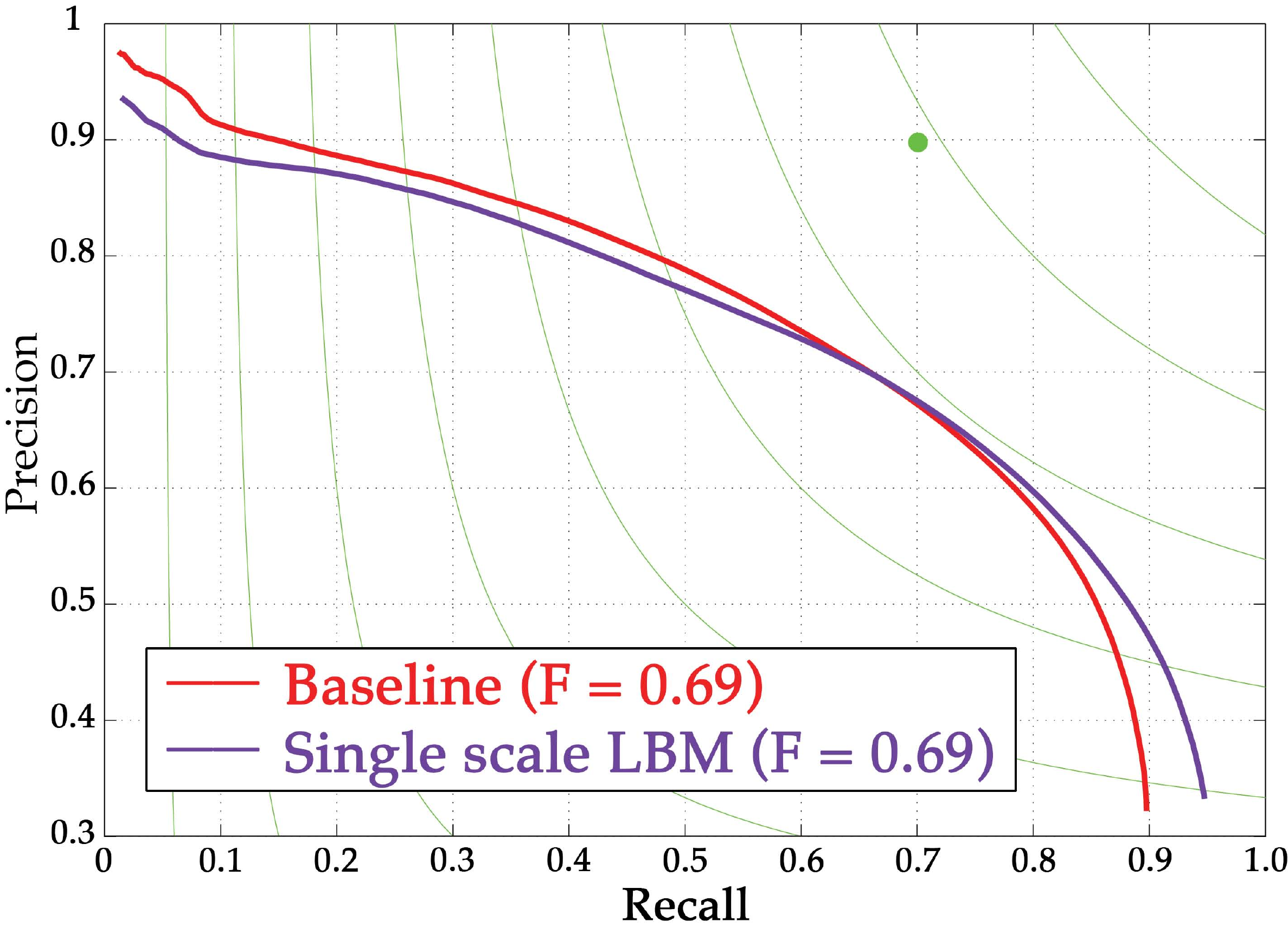}}
\subfigure[]{
\label{fig:pr:g}
\includegraphics[width=2.7cm]{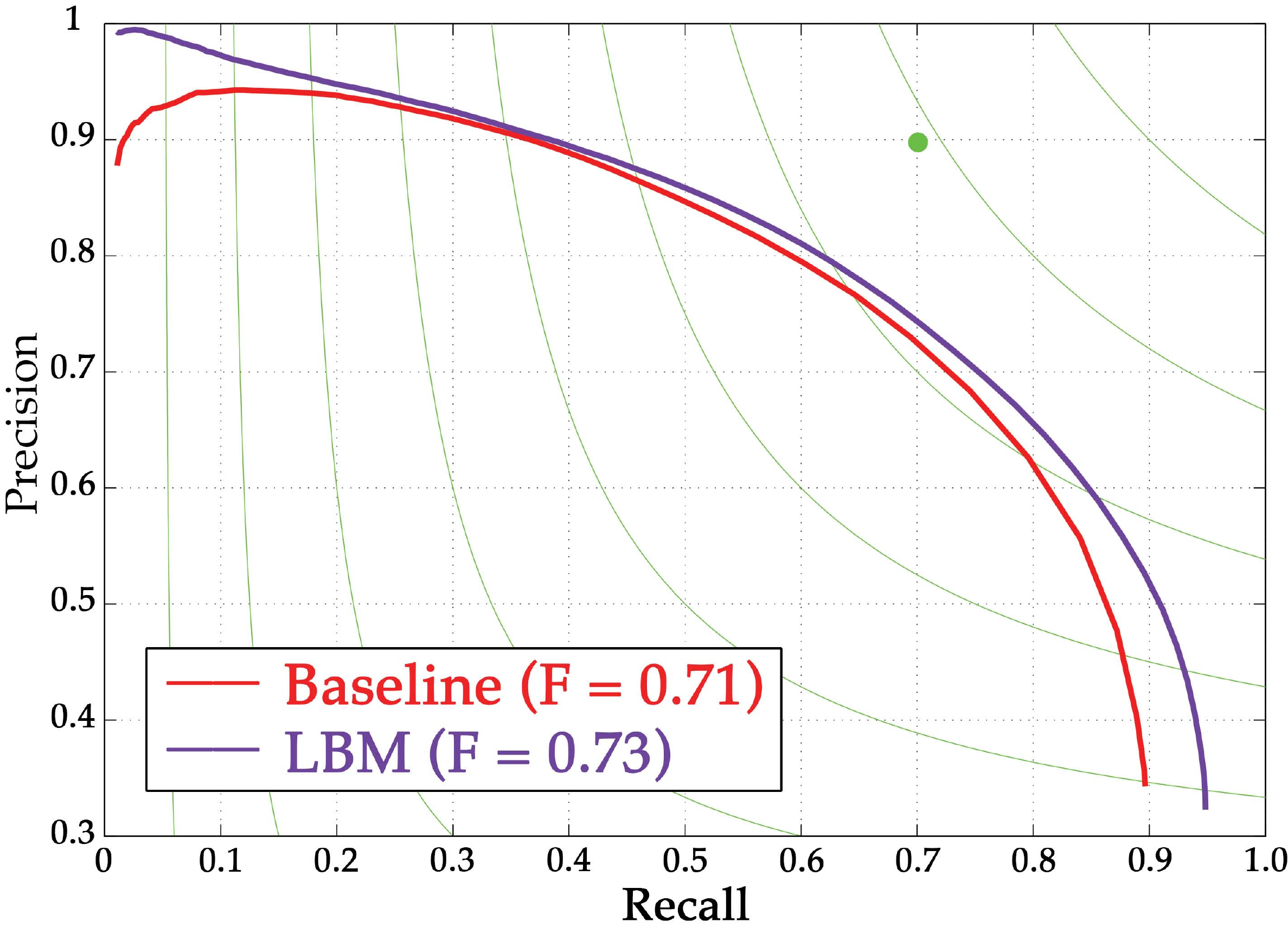}}
\caption{Precision-recall curves on BSDS500 benchmark.
(a) Comparison between $\chi^2$ difference and LBM using \emph{mPb} features.
After substituting $\chi^2$ difference with LBM,
the F-measure metric of \emph{mPb} is improved from 0.69 to 0.71.
(b) Comparison between $\chi^2$ difference and single scale LBM.
Even if only features at a single scale are available,
LBM achieves competitive results
compared with multi-scale approach of $\chi^2$ difference.
(c) Comparison between $\chi^2$ difference and LBM, both with globalization.
\emph{gPb} is the globalized version of \emph{mPb},
where an extra step of bootstrap is introduced.
After applying the same globalization method,
LBM also yields better results than \emph{gPb}.
All curves of LBM in this figure are results of LBM with RBF kernel.}
\label{fig:pr}
\end{figure}

\begin{figure*}[!t]
\centering
\includegraphics[width=18.0cm]{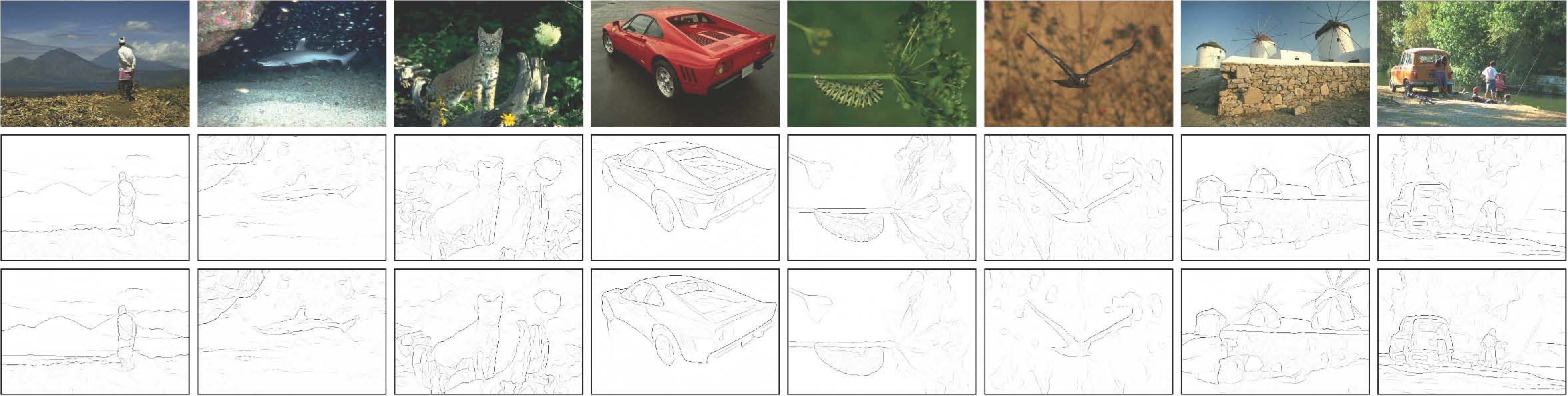}
\caption{Examples from the BSDS500 dataset.
Top row is source image, middle row is \emph{gPb} output,
and bottom row is LBM output with globalization (this work).
One advantage of our LBM approach is that
some hard boundaries are enhanced, such as the mountain and windmill.
Meanwhile, noisy boundaries of the red car, worm and owl are suppressed.}
\label{fig:ex}
\end{figure*}

Extensive experiments are conducted to verify the effectiveness of LBM.
Results are shown in Table~\ref{table:results},
Fig.~\ref{fig:pr} and Fig.~\ref{fig:ex}.
In Table~\ref{table:results}, ODS or OIS refers to the best F-measure
for the entire dataset or per image respectively,
and AP (Average Precision) is the area under the PR curve.
Apart from original images, noisy condition is also considered.
Here, we use Matlab R2012a to add Gaussian noise with default parameter.
To show the effectiveness of RBF kernel,
results of boundary metric using linear kernel
are presented in Table~\ref{table:results} as well.

According to results of experiment 1 and 2,
our algorithm compares favorably with the baseline approach,
for both original images and noisy ones.
After substituting $\chi^2$ difference with LBM,
the F-measure metric of \emph{mPb} is improved from 0.69 to 0.71.
The major advantage of LBM consists in the increase of maximum recall,
from 0.90 to 0.94 as shown in Fig.~\ref{fig:pr:m},
indicating that about 40\% of the missing pixels
of baseline approach are detected by LBM.
This results from the strong fitting capability of ANN,
which captures all kinds of variations of natural image data.
Experiment 3 only makes use of features at a single scale.
We find that the single scale LBM achieves competitive performance
compared with multi-scale approach of $\chi^2$ difference,
as shown in Fig.~\ref{fig:pr:s}.
Compared with the original \emph{mPb},
LBM learns more useful information from human annotations.
The effectiveness of the learning stage of LBM
can be confirmed by comparing the results
in Table~\ref{table:mpb} and Table~\ref{table:results}.

\begin{table}[!t]
\scriptsize
\caption{Comparisons between $\chi^2$ Difference and LBM}
\label{table:results}
\centering
\begin{tabular}{|cc|c|c|c|c|c|c|}
\hline
\multicolumn{2}{|c|}{\multirow{2}{*}{Method}} & \multicolumn{3}{|c|}{Original Image} & \multicolumn{3}{|c|}{Noisy Image} \\
\cline{3-8} & & ODS & OIS & AP & ODS & OIS & AP \\
\hline
1 & \emph{mPb} + $\chi^2$ difference & 0.69 & 0.71 & 0.68 & 0.67 & 0.68 & 0.67 \\
2 & \emph{mPb} + LBM (RBF) & \textbf{0.71} & \textbf{0.74} & 0.73 & \textbf{0.69} & \textbf{0.71} & \textbf{0.72} \\
3 & \emph{Pb} + LBM (RBF) & 0.69 & 0.71 & 0.70 & - & - & - \\
4 & \emph{mPb} + LBM (linear) & 0.70 & 0.73 & \textbf{0.74} & - & - & - \\
\hline
5 & \emph{gPb} + $\chi^2$ difference & 0.71 & 0.73 & 0.73 & 0.69 & 0.70 & 0.70 \\
6 & \emph{gPb} + LBM (RBF) & \textbf{0.73} & \textbf{0.75} & \textbf{0.78} & \textbf{0.71} & \textbf{0.72} & \textbf{0.76} \\
7 & \emph{gPb} + LBM (linear) & 0.72 & 0.74 & 0.77 & - & - & - \\
\hline
\end{tabular}
\end{table}

In \cite{arbelaez2011}, the authors introduce a globalization method
as a bootstrap to further improve the performance of \emph{mPb}.
The new algorithm is named as \emph{gPb}.
The proposed LBM can also be integrated into the framework of \emph{gPb}.
In the original work, boundary responses computed by the bootstrap step
is multiplied by a learned weight and added to \emph{mPb} output.
We follow a similar strategy, using the algorithm
introduced by \cite{jansche2005} to learn the weight.
According to experiment 5 and 6,
all 3 measurements of LBM produce better results than \emph{gPb}.
Corresponding PR curves can be found in Fig.~\ref{fig:pr:g}.
Apart from PR curves, standard deviation of best F-measures
for each image is also computed
to show the statistical significance of the improvement.
The standard deviation of $gPb$ + LBM (RBF) is $9.75 \times 10^{-3}$,
while that of $gPb$ + $\chi^2$ difference is $9.83 \times 10^{-3}$.
In addition, LBM obtains superior results in 131 out of 200 testing images.
Fig.~\ref{fig:ex} shows some examples.
One advantage of our LBM approach is that
some hard boundaries are enhanced, such as the mountain and windmill.
Meanwhile, noisy boundaries of the red car, worm and owl are suppressed.
What is more, these results are competitive
with the state-of-the-art results reported in \cite{ren2012}
(ODS: 0.74, OIS: 0.76 and AP: 0.77),
which take advantage of sparse coding based local features.

\section{Conclusion}
In this letter, a Learning-based Boundary Metric (LBM)
is proposed to substitute the $\chi^2$ difference used in \emph{mPb}.
One of the advantages of LBM is the strong fitting capability of natural image data.
With supervised learning, LBM is able to learn useful information from human annotations,
while the learning stage of \emph{mPb} achieves only limited improvements.
The structure of LBM is easy to understand,
composed of a single layer neural network and an RBF kernel.
With the above advantages, LBM yields better performance
than both \emph{mPb} and \emph{gPb}.
Extensive experiments are conducted to verify the effectiveness of LBM.
The F-measure metric on BSDS500 benchmark
is increased to 0.71 (without globalization)
and 0.73 (with globalization) respectively.
In the future, we are interested in applying LBM
to the framework of SCG, which achieves the state-of-the-art performance.

\bibliographystyle{IEEEtran}
\bibliography{LBM}

\begin{thebibliography}{10}
\providecommand{\url}[1]{#1}
\csname url@samestyle\endcsname
\providecommand{\newblock}{\relax}
\providecommand{\bibinfo}[2]{#2}
\providecommand{\BIBentrySTDinterwordspacing}{\spaceskip=0pt\relax}
\providecommand{\BIBentryALTinterwordstretchfactor}{4}
\providecommand{\BIBentryALTinterwordspacing}{\spaceskip=\fontdimen2\font plus
\BIBentryALTinterwordstretchfactor\fontdimen3\font minus
  \fontdimen4\font\relax}
\providecommand{\BIBforeignlanguage}[2]{{%
\expandafter\ifx\csname l@#1\endcsname\relax
\typeout{** WARNING: IEEEtran.bst: No hyphenation pattern has been}%
\typeout{** loaded for the language `#1'. Using the pattern for}%
\typeout{** the default language instead.}%
\else
\language=\csname l@#1\endcsname
\fi
#2}}
\providecommand{\BIBdecl}{\relax}
\BIBdecl

\bibitem{opelt2006}
A.~Opelt, A.~Pinz, and A.~Zisserman, ``A boundary-fragment-model for object
  detection,'' in \emph{ECCV}, 2006, pp. 575--588.

\bibitem{shotton2008}
J.~Shotton, A.~Blake, and R.~Cipolla, ``Multiscale categorical object
  recognition using contour fragments,'' \emph{IEEE Trans. Pattern Analysis and
  Machine Intelligence}, vol.~30, no.~7, pp. 1270--1281, 2008.

\bibitem{farhadi2009}
A.~Farhadi, I.~Endres, D.~Hoiem, and D.~Forsyth, ``Describing objects by their
  attributes,'' in \emph{CVPR}, 2009, pp. 1778--1785.

\bibitem{ferrari2010}
V.~Ferrari, F.~Jurie, and C.~Schmid, ``\BIBforeignlanguage{English}{From images
  to shape models for object detection},''
  \emph{\BIBforeignlanguage{English}{Int'l J. of Computer Vision}}, vol.~87,
  no.~3, pp. 284--303, 2010.

\bibitem{canny1986}
J.~Canny, ``A computational approach to edge detection,'' \emph{IEEE Trans.
  Pattern Analysis and Machine Intelligence}, vol.~8, no.~6, pp. 679--698,
  1986.

\bibitem{dollar2006}
P.~Dollar, Z.~Tu, and S.~Belongie, ``Supervised learning of edges and object
  boundaries,'' in \emph{CVPR}, 2006, pp. 1964--1971.

\bibitem{kokkinos2010}
I.~Kokkinos, ``Boundary detection using f-measure-, filter- and feature- (f3)
  boost,'' in \emph{ECCV}, 2010, pp. 650--663.

\bibitem{kennedy2011}
R.~Kennedy, J.~Gallier, and J.~Shi, ``Contour cut: Identifying salient contours
  in images by solving a hermitian eigenvalue problem,'' in \emph{CVPR}, 2011,
  pp. 2065--2072.

\bibitem{arbelaez2011}
P.~Arbelaez, M.~Maire, C.~Fowlkes, and J.~Malik, ``Contour detection and
  hierarchical image segmentation,'' \emph{IEEE Trans. Pattern Analysis and
  Machine Intelligence}, vol.~33, no.~5, pp. 898--916, 2011.

\bibitem{martin2004}
D.~Martin, C.~Fowlkes, and J.~Malik, ``Learning to detect natural image
  boundaries using local brightness, color, and texture cues,'' \emph{IEEE
  Trans. Pattern Analysis and Machine Intelligence}, vol.~26, no.~5, pp.
  530--549, 2004.

\bibitem{ren2008}
X.~Ren, ``Multi-scale improves boundary detection in natural images,'' in
  \emph{ECCV}, 2008, pp. 533--545.

\bibitem{martin2001}
D.~Martin, C.~Fowlkes, D.~Tal, and J.~Malik, ``A database of human segmented
  natural images and its application to evaluating segmentation algorithms and
  measuring ecological statistics,'' in \emph{ICCV}, 2001, pp. 416--423 vol.2.

\bibitem{eric2003}
E.~P. Xing, A.~Y. Ng, M.~I. Jordan, and S.~Russell, ``Distance metric learning
  with application to clustering with side-information,'' in \emph{NIPS}, 2003,
  pp. 505--512.

\bibitem{jansche2005}
M.~Jansche, ``Maximum expected f-measure training of logistic regression
  models,'' in \emph{Proceedings of the Conference on Human Language Technology
  and Empirical Methods in Natural Language Processing}, 2005, pp. 692--699.

\bibitem{ren2012}
X.~Ren and L.~Bo, ``Discriminatively trained sparse code gradients for contour
  detection,'' in \emph{NIPS}, 2012, pp. 593--601.

\end{thebibliography}

\end{document}